\documentclass[10pt,a4paper]{article}

\usepackage{cogsci}
\usepackage{makecell}

\cogscifinalcopy 

\usepackage{cmap}
\usepackage[T1]{fontenc}
\usepackage[american]{babel}
\usepackage{csquotes}
\usepackage{newtxtext,newtxmath}  
\usepackage{graphicx}
\usepackage{color,soul}
\usepackage[
  backend=biber,
  style=apa,
  natbib=true,
  annotation=false,
]{biblatex}
\usepackage{float} 

\usepackage[hidelinks]{hyperref}

\title{Modeling individual and item differences \\in atypicality inferences with ACT-R}

\author[2]{\mbox{Guifu Liu (guifu.liu@uni-saarland.de)}}
\author[1,2]{\mbox{Vera Demberg}}
\affil[1]{Max Planck Institute for Informatics}
\affil[2]{Saarland University}

\begin{document}

\maketitle

\begin{abstract}
Humans engage in pragmatic reasoning when encountering mentions of highly predictable events. Such mentions may appear redundant, leading comprehenders to infer that the event is atypical, and accommodate this inference with an explanation for why it is atypical. However, these inferences are not drawn uniformly across individuals, and experimental items vary in inference rates due to differences in event typicality. We show that these patterns can be predicted with a resource-rational cognitive model in which comprehenders select interpretation strategies based on individual exploration tendencies and item-level predispositions. These parameters are estimated independently from participants’ Raven’s Matrices score, event typicality rating and language model based accommodation likelihood. Taken together with recent findings from reference games, our results motivate a new perspective on the mechanisms that support pragmatic inferences.

\end{abstract}

\section{Introduction}

In language comprehension, humans go beyond recovering literal meaning by engaging in additional, context-sensitive reasoning to derive an  enriched interpretation. In this paper, we focus on atypicality inferences, pragmatic inferences that arise when comprehenders encounter information that is informationally redundant (IR). The redundancy arises when the information can be inferred from world knowledge about typical event sequences (or scripts, \cite{schank1975scripts}). 

 
\begin{quote}
    (1) Mary went to a restaurant. She ate there!
\end{quote}

Consider (1). A comprehender is likely to infer that going to a restaurant typically involves being seated, ordering, eating, and paying. As a result,  the utterance “She ate there!” is informationally redundant, since it expresses information that is already part of common-sense knowledge about everyday events and would normally be inferred automatically. Such redundancy violates the quantity maxim \citep{grice1975logic}, which should be avoided in rational communication.

However, 
mentioning the inferable event can lead to eliciting pragmatic inferences \citep{kravtchenko2015semantically}. A comprehender repairs the utterance's informativeness by inferring that the mentioned event is atypical for the referent. As in (1), a comprehender infers that Mary may be a less habitual eater than initially assumed, which justifies the need of an explicit mention (``She ate there!''). The comprehender may then accommodate atypicality by providing an alternative behavior or explanation with their world knowledge (e.g., Mary usually only orders drinks).

Nevertheless, humans do not draw atypicality inferences uniformly. \citet{ryzhova_what_2023} showed that the variability in subjects can be related to cognitive traits: individuals with higher reasoning abilities are more likely to draw atypicality inferences, and participants vary in how often they draw such inferences.  In addition to individual differences, experimental items vary in the typicality of the target events. For example, 
 \textit{picking the toppings} when ordering pizza is less typical than \textit{eating} when going to a restaurant.
Because atypicality inferences are context-sensitive, such item-level variation may further modulate inference rates.  

However, existing models have so far failed to account for individual and item differences when deriving atypicality inferences. Prior work using the Rational Speech Act (RSA) framework shows how such inferences can be derived in principle, but relies on a fully rational model that does not allow for individual differences \citep{kravtchenko_modeling_2022}. Similarly, \citet {kurch_large_2024} show that large language models fail to derive atypicality inferences, especially in adhering to the conversational norm that redundancies should be avoided.


Motivated by prior work modeling pragmatic inference in reference games \citep{DuffMaynDemberg2025}, we developed a resource-rational model of atypicality inferences in which comprehenders explore several possible interpretations of an informationally redundant utterance, trading off resource demands against expected success. We implement this model within the ACT-R cognitive architecture \citep{anderson2014atomic} and examine its behavior across a range of parameter settings: individual differences are captured by varying reinforcement learning parameters that modulate exploration tendencies \citep{stocco2021individual}, which serve as a proxy for comprehenders’ cognitive capacities in selecting interpretation strategies. Item-level differences are modeled by adjusting the space of possible interpretations with their respective utility, allowing a  realistic account of belief updates when deriving inferences.



We first show in two sensitivity analyses that the model can generate patterns of individual and item differences when varying these parameters.  Given that prior work highlights the effects of reinforcement learning parameters in an ACT-R model of Raven’s Matrices \citep{raven1962advanced}, our result provides a plausible explanation for the correlation between atypicality inferences and Raven’s performance observed in \citet{ryzhova_what_2023}. We then independently estimate these parameters to ground such sources of variation: individual differences are informed by Raven’s performance from \citet{ryzhova_what_2023}, while item differences are derived from language model surprisal for accommodations and from human habituality ratings in \citet{kravtchenko_informationally_2022}. We find that, even without task-specific training, the model improves the prediction of inference performance across participants. 

\begin{table}[t]
\centering
\caption{An example of \textit{restaurant} story with IR condition from \citet{ryzhova_what_2023}. The IR utterance is in bold.}
\vskip 0.12in
\label{task_example}
\resizebox{!}{0.12\textheight}{%
\begin{tabular}{p{1.6cm}p{6.8cm}}
\hline
\textbf{Context}              & Mary is a journalist who often goes to restaurants after her interviews. Yesterday she went to a popular Chinese place where she ran into her friend David. Later that day David ran into Sally, a mutual friend of him and Mary. \\
\hline
\textbf{\makecell[tl]{Target\\event}}     & David said to Sally: ``I ran into Mary  leaving that Chinese place. \textbf{She ate there!}''                                                                                                                                                \\
\hline
\textbf{Question 1}           & \textit{How often do you think Mary usually eats,  when going to a restaurant?}                                                                                                                                                            \\
                     & \textbf{Answer:} 82                                                                                                                                                                                                                                \\
                     \hline
\textbf{Question 2}           & \textit{Why did you think it is this often?}                                                                                                                                                                                               \\
                     & \textbf{Answer:} Going to a restaurant normally involves eating, although one could go for just a drink.                                                                                                                                           \\
                     \hline
\textbf{\makecell[l]{Inference\\annotation}} & \textit{atypicality} \\ \hline                                                                                                                                                                                                                      
\end{tabular}
}

\end{table}

\section{Background}
\subsubsection{Atypicality Inferences.}We adapt the experimental procedure of \citet{ryzhova_what_2023} for our atypicality inference task. The materials consist of twenty-four brief stories describing everyday situations such as going to a restaurant (see an example in \autoref{task_example}). Each item consists of a context of a few sentences that introduces the topic and story characters. To introduce information redundancy, the story continues with an utterance where a character states an event that is highly predictable given the context (“She ate there!”). 

Participants first rate the typicality of the target event on a scale of 0 to 100 (Question 1) and then provide a brief textual explanation for their rating (Question 2). The explanations are annotated as follows to determine whether a low typicality rating truly reflects an atypicality inference triggered by the redundant utterance: \textit{normal} (no inference), \textit{atypicality} (inference correctly drawn), \textit{notice reject} (made the inference but did not accept), \textit{not sure} (participant expresses uncertainty) and \textit{other} (responses not fitting the above categories). Example responses for each label are shown in \autoref{annotation_example}. Some responses contain accommodations, or explanations for why the target event is atypical (shown in bold in \autoref{annotation_example}). Each participant is shown only 6 items to avoid learning effects. 

In our experiments, we simplify the two-step procedure by directly eliciting interpretation strategies that correspond to these annotation categories (see \hyperref[actr_model]{the next section}). Compared to ratings, annotations are less noisy in representing what inferences are actually made \citep{ryzhova_what_2023}.

\begin{table}[t]
\centering
\caption{Examples of annotations from \textit{restaurant} story from \citet{ryzhova_what_2023}. The accommodation is in bold.}
\vskip 0.12in
\label{annotation_example}
\resizebox{!}{0.12\textheight}{%
\begin{tabular}{p{1.7cm}p{1.5cm}p{5cm}}
\hline
\textbf{\makecell[tl]{Inference\\Annotation}} & \textbf{\makecell[tl]{Inference\\Drawn}} & \textbf{Example}                                                                                                                             \\ \hline
\textit{normal}               & no              & Usually when you go to a restaurant, it is to eat.                                                                                  \\ \hline
\textit{atypicality}          & yes             & Going to a restaurant normally involves eating, although one could go \textbf{for just a drink}. \\ \hline
\textit{\makecell[tl]{notice\\reject}}        & unclear         & After interviews Mary will be tired so she probably eats. She can’t just go to a restaurant \textbf{for a drink} after a long day.           \\ \hline
\textit{not sure}             & unclear         & I’m not sure.                                                                                                                       \\ \hline
\textit{other}                & unclear         & He didn’t tell Sally which restaurant, he said that restaurant, as though they go there often. \\ \hline                               
\end{tabular}
}

\end{table}

\subsubsection{Individual Differences.} 
Comprehenders interpret redundant utterances differently in the atypicality inference task. One source of this variation is general problem-solving ability measured by Raven’s Progressive Matrices \citep{raven1962advanced}, which present an incomplete grid of geometric shapes that a participant must complete by identifying the underlying pattern. Higher scores in Raven’s matrices are linked to higher rates in both atypicality inferences \citep{ryzhova_what_2023} and target selection in reference games \citep{mayn2022individual}, another pragmatic task with particularized implicatures.

Prior work offers some insight into the cognitive mechanism underlying performance on Raven’s matrices and its connection with pragmatic reasoning. Stocco et al. (2021) proposed that individual differences in Raven’s performance arise from differences in how participants explore the hypothesis space of possible solutions to the task --- specifically, in persistence (how long they explore possible hypotheses before giving up) and strength of negative feedback (how quickly they disengage from failed hypotheses). These mechanisms, supported by reinforcement learning, were implemented in ACT-R and validated through a secondary task and an fMRI follow-up. \citet{DuffMaynDemberg2025} subsequently applied this account to a pragmatic reference game, where variation in the strength of negative feedback is also found to correlate with task performance.




\subsubsection{Item Differences.}  \citet{kravtchenko_informationally_2022} discussed differences in habituality across items. The distinction between highly predictable and less predictable events is gradual, ranging from very likely events ($p > 0.9$) to those only slightly more probable than the non-habitual ones.  Differences in pre-utterance belief also affect atypicality inferences: relatively more habitual events tend to elicit more atypicality inferences. In addition, item difference in inference performance is also found in \citet{ryzhova_what_2023}, where per-item atypicality inference rate (trials annotated as \textit{atypical}) varies from $15.38\%$ to $71.42\%$. We hypothesize that models that take target event and accommodation typicality as item properties can better predict inference performance.



\section{ACT-R Model} 
\label{actr_model}

\begin{figure}
\centering
 \includegraphics[width=0.87\linewidth]{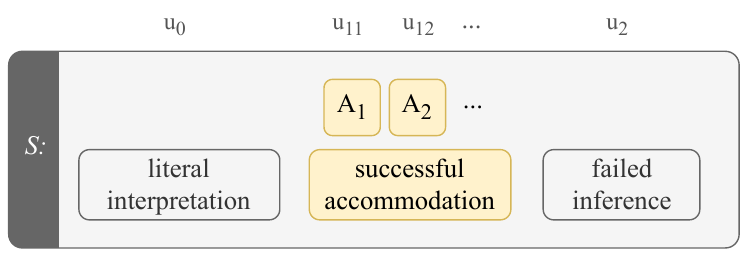}
  \caption{The set of interpretation strategies $S$ and their respective utility $U$.}
  \label{schematic_b}
\end{figure}
\begin{figure}[t]
  
  \centering
    \includegraphics[width=0.87\linewidth]{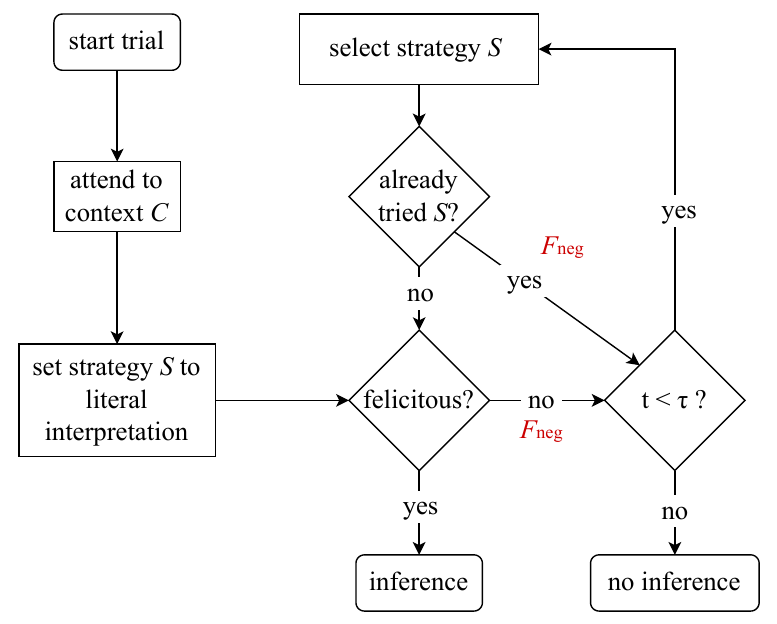}
  
  \caption{An overview of our model’s control flow adapted from \citet{duff_role_2026}. $S$ is a strategy in reaching an interpretation, $A$ is a distinct accommodation for the atypical action, $t$ is the model runtime, and $\tau$ is persistence. Any \textit{successful accommodation} strategy is felicitous.}
  \label{schematic_a}
\end{figure}
We create a computational model of a comprehender performing an atypicality inference task. The model has three requirements: (1) it implements several interpretation strategies for the task under the constraints of human cognition, (2) it can be parameterized individually to examine how individual differences in cognitive capacity may impact task performance, and (3) it can be parameterized by items to examine how differences in habituality prior of the target event and likelihood estimates of accommodations may impact task performance. To satisfy these requirements, we choose to specify our model in the ACT-R cognitive architecture \citep{anderson2014atomic}. ACT-R provides a framework in which a comprehender can explore alternative interpretation strategies under constraints of time and cognitive resources, modulated by individual-level exploration tendencies and item-level predispositions. Strategy selection is guided by efficiency through temporal-difference learning  \citep{fu2006recurrent}, by penalizing strategies that lead to success only after a long delay. We implement the model using the \texttt{pyactr} library \citep{brasoveanu2020computational}.

\subsubsection{Design.} 
The comprehender can adopt one of three strategies when interpreting an utterance: literal interpretation (no atypicality inference is drawn), successful accommodation (an atypicality inference is drawn and accommodated), and failed inference, see \autoref{schematic_b}. The latter includes cases in which the accommodation attempt fails to produce a coherent interpretation. Utilities $u$ encode both the likelihood of success and the cost to carry out that strategy. 

\autoref{schematic_a} illustrates the derivation of atypicality inferences in ACT-R. Each trial begins with an attempt to interpret an informationally redundant utterance (e.g., ``She ate there!'') literally.\footnote{This corresponds to the resource-rational adaptation strategy in \citet{duff_role_2026}, where a comprehender prefers to use simpler literal interpretation strategies until experience motivates the use of additional resources. A richer inference of reconciling the atypical event with world knowledge isn't readily available and should incur processing cost \citep{ryzhova_processing_2023}.} 
An interpretation is pragmatically felicitous if it does not violate Gricean maxims \citep{grice1975logic}. This condition is not met for informationally redundant utterances under a literal interpretation. Consequently, the utility of the literal interpretation is updated using negative feedback $F_\text{neg}$, which lowers its expected utility. 
If runtime remains below persistence ($t < \tau$), the model selects a strategy $S$ based on their expected utility $u$. The strategy with the highest utility is selected, subject to utility noise. If the selected strategy has been tried before, its utility is further reduced by the negative feedback $F_\text{neg}$. Otherwise, the new interpretation is evaluated for pragmatic felicity. If it is felicitous, the model terminates with an atypicality inference. If not, the process iterates until either a felicitous interpretation is found or the runtime exceeds persistence. In the latter case, no atypicality inference is drawn.



\subsubsection{Evaluation.} The experimental items originate from \citet{kravtchenko_informationally_2022}, but here we use \citet{ryzhova_what_2023} as the test data, which we introduced in the previous section. They additionally elicited and categorized participants' inferences. We focus specifically on model performance in the critical trials, which contain the informationally redundant utterance (see \autoref{task_example}). We note though that our model can also handle non-critical trials, in which the comprehender trivially adopts a literal interpretation that is pragmatically felicitous. 

We evaluate the model on annotated participant responses, using annotation tags that indicate whether and how an atypicality inference was made. 
Responses labeled \textit{not sure} or \textit{other} are excluded, as they do not clearly reflect an inference (8.8\% of trials). The resulting dataset consists of 1,056 responses across 24 story items and 193 participants. Each response falls into one of three annotation categories: \textit{normal} (43.3\%), \textit{atypicality} (50\%), and \textit{notice reject} (6.7\%). See the previous section for definitions and examples of each label.


\section{Sensitivity Analyses}
We first examine how the model’s predictions change as we vary persistence and negative feedback strength $F_\text{neg}$ independently. We hypothesize that these factors account for variance in the atypicality inference task, given that they have been shown to explain variance in Raven’s Matrices \citep{stocco2021individual}, which in turn correlates with atypicality inference performance \citep{ryzhova_what_2023}. We also investigate how the model's predictions change as we vary the initial expected utility of interpretation strategies. We hypothesize that the model can also capture hypothetical differences in predisposition towards pragmatic reasoning.  

\subsubsection{Method.} We conduct two analyses to examine whether (1) $F_\text{neg}$ and persistence and (2) the initial expected utilities reliably control accuracy in the simulated data. Based on pilot simulations, in \textbf{Analysis 1} we first examine performance with one of twenty $F_\text{neg}$ values between $0.05$ and $10$, and one of twenty persistence values between 25 and 45  (evenly spaced). Initial utilities for literal $u_0$, successful $u_1$ and failed accommodation $u_2$ are fixed to $5, -2.5, -5$, respectively.\footnote{The utility value can be interpreted in conjunction with the success rate of an interpretation strategy in achieving the inference and the cost, or temporal offset between strategy selection and the reward.} We run 5 simulations for each combination of values, which results in a total of 2,000 simulations. In each simulation, we follow the experimental lists in \citet{ryzhova_what_2023}, with 6 observations per subject for a total of 1,158 observations. 

In \textbf{Analysis 2}, we vary the possible values for the initial expected utility for literal interpretation $u_0 \in \{5, 2.5, 0\}$ and successful accommodation $u_1 \in \{2.5, 0, -2.5, -5\}$. We hold the starting utility of failed inference at $-5$. As we assume comprehenders should always prefer literal interpretation initially, we investigate combinations of values where $u_0 > u_1$. We choose five $F_\text{neg}$ values between $0.05$ and $10$ and five persistence values between 25 and 45. We run 5 simulations for each combination of values, which result in a total of 1,125 simulations. 

In both analyses, we use the following parameters for utility learning: the default learning rate of 0.2 and noise 0.6. Other model parameters are set to the default values and best practices for ACT-R model \citep{actr_tutorial_2022}.

\begin{figure}[t]
  \begin{center}    \includegraphics[width=1.\linewidth]{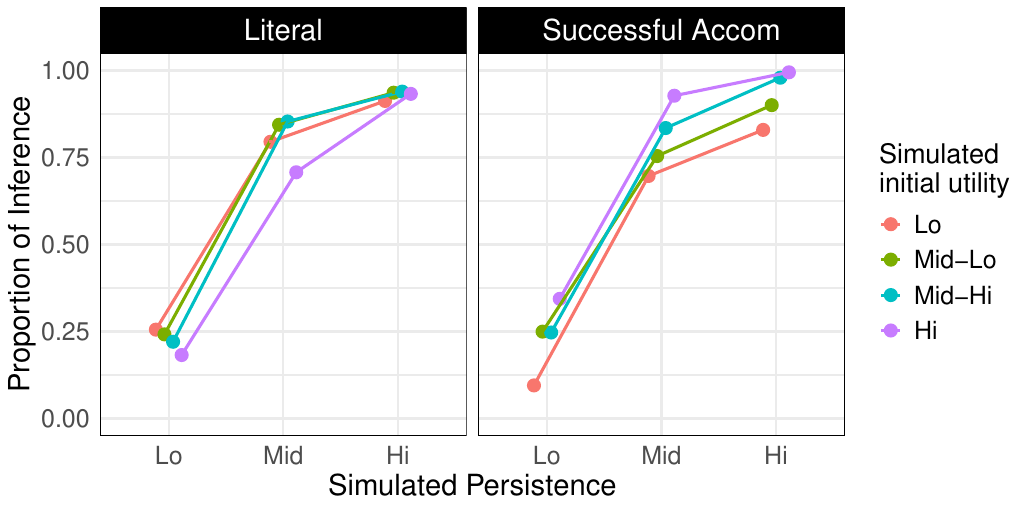}
  \end{center}
  \caption{The effect of reinforcement learning parameters, persistence and initial utilities (Left: $u_0$, Right: $u_1$), on inference rate in \textbf{Analysis 2}.}
  \label{experiment1a}
\end{figure}
\subsubsection{Results and Discussion.} 

We fit a Bayesian logistic regression to both sets of simulated responses, using the R package \texttt{brms}. We report $95\%$ credible-interval level for $\hat{\beta}$.
In \textbf{Analysis 1}, where we vary only $F_\text{neg}$ and persistence, our model finds positive relationships between inference accuracy and both $F_\text{neg}$, $\hat{\beta} = 1.56, [1.55, 1.56]$, and persistence, $\hat{\beta} = 3.63, [3.62, 3.64]$. An additional joint interaction indicates that $F_\text{neg}$ has its largest effect (on the log odds scale) at higher values of persistence (and vice versa), $\hat{\beta} = 0.95, [0.94, 0.96]$. 

In \textbf{Analysis 2}, we observe a negative effect of $u_0$, $\hat{\beta} = -0.67, [-0.68,-0.66]$ and a positive effect of $u_1$, $\hat{\beta} = 2.35, [2.34, 2.37]$. The results indicate that a high predisposition of literal interpretation can hinder inference while a high predisposition of pragmatic strategy helps an inference to be made. In addition, persistence is shown to have a stronger effect on inference performance than initial expected utility (\autoref{experiment1a}). 

Given that persistence and negative feedback account for variance in Raven's Matrices \citep{stocco2021individual}, our model provides a proof of concept to explain the positive relationship between Raven's performance and atypicality inference in \citet{ryzhova_what_2023}. 
The analyses demonstrate that the proposed model can reproduce the observed variation in inference across items and participants. In the following experiment, we test whether the model can predict an individual’s inference performance on a given item using independent estimates derived from that participant’s Raven’s Matrices performance and item properties. The procedure for parameter estimation is described in the next section.

\section{Modeling individual and item differences}

\subsubsection{Individual Differences} We use per-participant scores of Raven’s Progressive Matrices Test from \citet{ryzhova_what_2023}, which is approximately normally distributed ($\mu = 5.38, \sigma = 2.12$).   We map them to persistence $\tau \in [25, 45]$ as we find it has a larger effect on inference performance than $F_\text{neg}$ in the previous analysis. 
In \citet{stocco2021individual}, persistence has an effect on performance in Raven’s Matrices. In addition, participants with higher scores on Raven's tend to give more pragmatic responses for atypicality inference task \citep{ryzhova_what_2023}. We thereby hypothesize that models that incorporate persistence estimated by Raven's score can better predict inference performance. In such model, a pragmatic comprehender will take longer effort in exploring interpretation strategies for the redundant utterance. 




\subsubsection{Item Differences}
In addition to individual differences, experimental items vary in the typicality of the target events, which can in turn modulate inference rates. We use per-item initial utility of literal and pragmatic strategies to model item differences, which are estimated with human habituality ratings and language model surprisal.  

Atypicality inferences are highly context-sensitive and have been shown to disappear in atypical contexts (e.g., Mary does not like going to restaurants). We thereby hypothesize that the differences of context typicality across items can affect inference performance. As in \citet{kravtchenko_modeling_2022}, we estimate a prior for the habituality of the events with a beta distribution. Specifically, we use the pre-utterance habituality rating (i.e.~how habitual events are believed to be based on the context alone) collected in \citet{kravtchenko_informationally_2022}. The resulting prior estimate ranges from $0.65$ to $0.91$. The percentile rank of prior estimate in each item is mapped to \textbf{literal utility} $u_0 \in [1.0,5.0]$. 

Beyond literal interpretation, inference depends on the comprehender’s belief in possible \textbf{accommodations}: explanations that make an otherwise redundant utterance informative (e.g., “She doesn’t usually eat because she goes for a drink”). We generate five accommodations using GPT-5.2, and in a separate setting, we extend them with accommodations elicited from human participants in \citet{ryzhova_what_2023}.
All accommodations are grouped by semantic equivalence. 

For each accommodation, we generate 20 paraphrases using GPT-5.2 to expand its lexical realization. Belief strength is estimated by the surprisal of accommodation under GPT-J-6B \citep{gpt-j}, conditioned on the preceding context (e.g., the context and target event in \autoref{task_example}). GPT-J-6B is used because its surprisal closely aligns with N400 responses, a neural marker of semantic processing  \citep{michaelov2022more}. Language model surprisal is also shown to reliably reflect semantic plausibility \citep{kauf_log_2024}.\footnote{We also ask language models to directly judge how surprising an accommodation is in context (\cite{chhun2022human,deering2025analysis}), but this underperforms the surprisal-based approach.} 
This approach is less costly than collecting human ratings and provides item-level estimates that are not affected by individual variability, which we do not model here.

The paraphrase with the lowest surprisal, or the most probable, is selected, and its surprisal is mapped to an \textbf{accommodation utility} $u_{1} \in [-5.0, 1.0]$\footnote{ The probability of selecting a strategy with utility $u_i$ is $\frac{e^{\frac{u_i}{t}}}{\sum_{j=1}^n e^{\frac{u_j}{t}}}$}. Negative utilities indicate that accommodations are not helpful in language comprehension outside of information redundancy and ensure that literal interpretation needs to be penalized substantially before the model explores accommodations. 
Accommodation order is randomly shuffled across runs to avoid order effects.

\begin{table*}[t]
\centering
\caption{Overall and per-51.8 performance of Experiment 2. The results are averaged over 25 runs.}
\label{experiment2}
\vskip 0.12in
\resizebox{0.9\textwidth}{!}{%

    \begin{tabular}{l|ccc|cccc|ccc}
    \hline
    \textbf{Model} & \textbf{Indiv} & \multicolumn{2}{c|}{\textbf{Item}} 
    & \textbf{Acc.} & \textbf{F1} & \textbf{Sens.} & \textbf{Spec.} 
    & \textbf{Atypicality} 
    & \textbf{Notice Reject} & \textbf{No Inference} \\
    
     & Persis. & Accom & Literal 
     &  &  &  &  &  &  & \\
    \hline
    
    Baseline &  &  &  
     & 51.81$\pm$0.52 & 59.25 & 61.78 & 38.76 
     & 61.73 & 62.08 & 38.76 \\ 
     Indiv & \checkmark &   &  
     & 56.38$\pm$0.30 & 60.63 & 59.19 & \textbf{52.73} 
     & 58.67 & 63.04 & \textbf{52.73} \\
    Item &  & \checkmark & \checkmark  
     & 55.34$\pm$0.34 & \textbf{68.54} & \textbf{85.79} & 15.42 
    & \textbf{85.90} & \textbf{84.96} & 15.42 \\
    

     Indiv + Item & \checkmark &  & \checkmark
     & 56.73$\pm$0.26 & 62.35 & 63.18 & 48.27 
     & 62.72 & 66.59 & 48.27 \\
    

     & \checkmark & \checkmark &  
     & 56.81$\pm$0.31 & 62.67 & 63.91 & 47.53 
     & 63.14 & 69.58 & 47.53 \\
    
     & \checkmark & \checkmark & \checkmark  & \textbf{57.31$\pm$0.32} & 64.29 & 67.77 & 43.57 
     & 67.20 & 72.00 & 43.57 \\
    \hline
    \end{tabular}
    }

\end{table*}

\section{Model Experiment}



We test whether a model with parameters independently estimated from participant reasoning capability and item properties can better predict inference performance.

\subsubsection{Method.}  We estimate parameters of individual and item differences by following the procedures in the previous section. All other parameters are held the same as in previous analyses.   

To evaluate the quality of our model, we test whether it outperforms simpler alternatives in predicting inference. We compare four models: (1) \textbf{Baseline}, a fully fixed model using the best-performing parameters (i.e. persistence, $F_\text{neg}$ and initial expected utilities) from previous sensitivity analyses; (2) \textbf{Indiv}, a model in which initial utilities are fixed to the same values as the baseline. It assumes no item-specific predispositions in interpretation strategies, while allowing persistence to vary across individuals; (3) \textbf{Item}, a model in which persistence and $F_\text{neg}$ are fixed to the same values as the baseline but allow initial expected utilities to vary across items. (4) \textbf{Indiv + Item}, the full model and its ablations where the initial expected utilities for either literal interpretation or accommodations are fixed to the same value as the baseline. For each model, the results are averaged over 25 runs.

Recall that our test data are annotated for whether an atypicality inference is made. Trials labeled \textit{normal} are treated as cases in which no inference is drawn, whereas \textit{atypical} and \textit{notice reject} indicate that an inference is made. \textit{Notice reject} comprehenders are pragmatic and closer to \textit{atypical} than \textit{normal} comprehenders, as they elicit an accommodation (e.g. \textit{go for a drink} in \autoref{annotation_example}). A model’s prediction is considered correct if it matches the annotation regarding whether an inference occurred. Our models don't have free parameters as they are either fixed or estimated from other tasks. We report overall and per-category accuracy (which measures how well the model predicts inference performance for each annotation category). We also report \textit{sensitivity} (how well the model identifies instances in which an individual makes an atypical inference)  and \textit{specificity} (how well the model identifies cases in which an individual does not make an atypical inference).

\subsubsection{Results and Discussion.} \autoref{experiment2} presents overall and per-category performance. The results show that by incorporating individual and item differences with parameters fitted from other tasks, the model better predicts inference performance across participants. Both \textbf{Indiv} and \textbf{Indiv + Item} outperform the fully fixed baseline in overall metrics. The full model achieves the best performance in accuracy, although the \textbf{Indiv} model surpasses the other models in specificity.

We find only a small improvement of the full model over \textbf{Indiv} when item-level differences are added. We believe this is partly because variation across individuals has a greater impact on inference performance than variation across items. Differences in inference rates across items are relatively gradual, with most items centered around a 50\% inference rate. In contrast, inference rates vary much more widely across individuals. This is consistent with prior work showing that the effect of by-item pre-utterance beliefs on atypicality inference is gradient rather than categorical \citep{kravtchenko_informationally_2022}. In addition, the full \textbf{Item} model
under-performs \textbf{Indiv} model in overall accuracy, as it consistently overpredicts inference in all items partly due to lower initial literal utility. In previous sensitivity analyses, we also observe stronger effects of persistence, which modulate individual-level exploration tendencies, on inference performance than initial expected utility, which modulates item-level predispositions (\autoref{experiment1a}).  

\begin{figure}[t]
    \centering
    \includegraphics[width=0.6\linewidth]{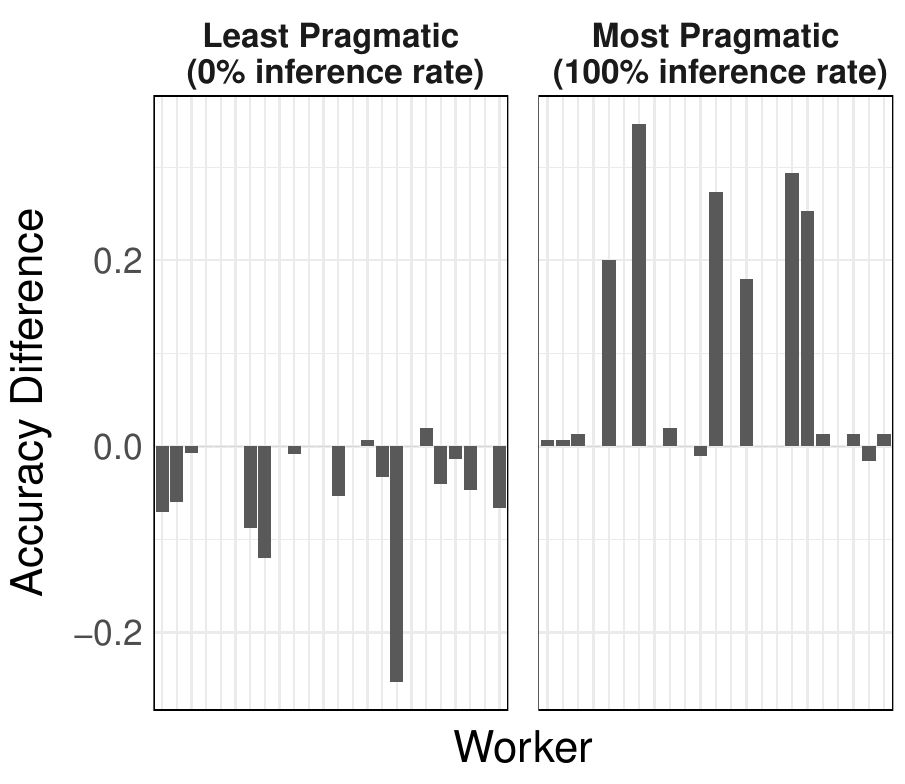}
    \caption{Per-participant accuracy difference between the full \textbf{Indiv + Item} and \textbf{Indiv} model for least and most pragmatic participants (by their atypicality rates in \cite{ryzhova_what_2023}).}
    \label{model_comparison}
\end{figure}

We also observe a tradeoff between sensitivity and specificity between \textbf{Indiv} and the full \textbf{Indiv + Item} model. Notably, adding item differences helps detect inferences ($+5.15\%$) but hurts accuracy in non-inferences ($-4.94\%$). The \textbf{Indiv} model better captures non-inference especially when comprehenders consistently make no inference, as item-level variation provides little help (\autoref{model_comparison}). In contrast, pragmatic comprehenders can possibly discern fine-grained item differences, allowing the full model to detect their inference better. 

One potential concern is that our distribution of accommodations may not meaningfully affect per-item inference performance. However, our results indicate otherwise. We observe that several items received less than five distinct accommodations derived from humans and language model, which should in turn lower the predicted inference rate. Indeed, in \citet{ryzhova_what_2023}, two items exhibit the lowest inference rates (15.38\% for \textit{exhibit} and 29.27\% for \textit{baking}), while two others (41.17\% for \textit{shampoo} and \textit{bus}) fall below a majority. 

In addition to accommodations elicited by both humans and the language model, we evaluate those elicited solely by the language model. The resulting accuracy for the full \textbf{Indiv + Item} model is comparable at 57.02 $\pm$ 0.24, which is expected given that model-elicited accommodations closely align with human ones. Each item has at least two accommodations elicited by language model that are semantically equivalent to those provided by humans.

\section{General Discussion} 
In this work, we introduced a mechanistic, resource-rational model of atypicality inference derivation and showed that variation in performance can be explained by participant-level exploration tendencies and item-level predispositions. This result provides further evidence that such factors contribute to individual differences in pragmatic tasks besides reference games \citep{DuffMaynDemberg2025}. We further showed that, even without task-specific training, reinforcement-learning parameters estimated from correlated measures can predict individual inference behavior. This finding supports earlier claims linking atypicality inference to cognitive traits such as Raven’s Matrices performance \citep{ryzhova_what_2023} and habituality ratings \citep{kravtchenko_informationally_2022, kravtchenko_modeling_2022}. 

Our approach has several limitations. First, the current model lacks a mechanism for rejecting otherwise felicitous accommodations (i.e., \textit{notice reject}). This can possibly be addressed by adding an evaluative step after felicity checking, in which a candidate strategy is subject to the comprehender’s beliefs about how plausible is the accommodated event  (e.g., She can't just go to a restaurant for a drink). Moreover, the model experiment shows that the model cannot dynamically adjust item-level predispositions across individuals with different pragmatic profiles. This suggests a more complex interplay between reasoning ability, pragmatic sensitivity, and individual experience than the model currently captures. Factors such as creativity in generating explanations and personal experience with the target events likely play an important role. Since we don't have participant-level estimates of accommodations, the use of language model may have created imbalance between \textbf{Individual} and \textbf{Item} predictors, especially in that it does not account for human failures in accommodation.

Future work should collect more fine-grained behavioral measures to better validate the proposed framework. In particular, measures such as the strength of negative feedback ($F_\text{neg}$) and persistence in secondary tasks may help clarify their roles in atypicality inference, as a similar approach has been attested on reference games and Raven’s Matrices \citep{duff_role_2026}. Finally, because explanations in \citet{ryzhova_what_2023} were collected post hoc, it remains an open question whether comprehenders engage in online inference derivations comparable to those assumed by our model. Addressing these issues will be crucial for refining computational accounts of individual and item differences in pragmatic inference.

\section{Acknowledgments}
This work is funded by the Deutsche Forschungsgemeinschaft (DFG, German Research Foundation) -- GRK 2853/1 “Neuroexplicit Models of Language, Vision, and Action” - project number 471607914.

\printbibliography

\end{document}